\documentclass[nohyperref]{article}

\usepackage{microtype}
\usepackage{graphicx}
\usepackage{subfigure}
\usepackage{booktabs} 

\usepackage{hyperref}

\usepackage[accepted]{icml2022}

\usepackage{amsmath}
\usepackage{amssymb}
\usepackage{mathtools}
\usepackage{amsthm}

\usepackage[capitalize,noabbrev]{cleveref}

\theoremstyle{plain}

\theoremstyle{definition}

\theoremstyle{remark}

\usepackage[textsize=tiny]{todonotes}

\icmltitlerunning{Transformer-Based Generative Engine Optimization}

\begin{document}

\twocolumn[
\icmltitle{Beyond SEO: A Transformer-Based Approach for \\ Reinventing Web Content Optimisation}



\icmlsetsymbol{equal}{*}

\begin{icmlauthorlist}
\icmlauthor{Florian Lüttgenau,}{yyy}
\icmlauthor{Imar Colic}{xxx}
\icmlauthor{Gervasio Ramirez}{xxx}
\end{icmlauthorlist}

\icmlaffiliation{yyy}{Department of Management, London School of Economics and Political Science, London, United Kingdom}
\icmlaffiliation{xxx}{Department of Statistics, London School of Economics and Political Science, London, United Kingdom}

\icmlcorrespondingauthor{Florian Lüttgenau}{f.e.luttgenau@lse.ac.uk}
\icmlcorrespondingauthor{Imar Colic}{i.colic@lse.ac.uk}

\icmlkeywords{GEO, Machine Learning, BART, Deep Learning, Search Optimization}

\vskip 0.3in
]



\printAffiliationsAndNotice{} 

\begin{abstract}
This paper presents a fine-tuned BART-base model for optimising travel website content to improve visibility in generative search engines. Using synthetically generated training data of 1905 cleaned instances and domain-specific strategies, the model was trained in a supervised approach and achieved measurable gains in ROUGE-L and BLEU compared to the BART-base model. In simulations, our proposed model boosted website visibility in generative engines by up to 30.96\%. Our work shows that even small-scale, domain-focused fine-tuning can significantly enhance content discoverability in AI-driven search engines.
\end{abstract}

\section{Introduction}
\label{submission}
The era of keyword-driven search engines pioneered and lead by Google, and conventional SEO (Search Engine Optimisation) tactics focused on keyword placement and backlink building is coming to an end. This disruption stems from the rise of large language models (LLMs) and the rapid adoption of generative AI-driven search models powered such as ChatGPT Search or Perplexity AI \citep{amer2024}. Unlike conventional search engines that return ranked link lists, generative engines draw on multiple sources to craft coherent, conversational, and personalized answers—transforming how information is retrieved, how users interact with given content, and how content and information is discovered \citep{aggarwal2024}.

By 2026, Gartner predicts a 25\% reduction in conventional search engine usage \citep{gartner2024}, signalling a paradigm shift that poses substantial challenges for website owners, digital marketers, and content creators striving to maintain online visibility. The opaque, fast-changing nature of generative models intensifies this challenge, underscoring the urgent need for fresh theoretical frameworks and practical strategies.   

Recent work by \citet{aggarwal2024} proposes Generative Engine Optimisation (GEO), a new framework that rethinks visibility as a multidimensional metric rooted in how web content is transformed and evaluated within generative model outputs. Leveraging generative pre-trained transformers (GPTs), their approach uses a black-box optimisation pipeline and benchmarks multiple optimisation strategies across domains. GEO uses a set of optimisation strategies that differ from conventional SEO practices, like for instance: Content fluency, quotation addition, statistical data, or authoritative tone. \citet{aggarwal2024} demonstrate that previous SEO techniques are largely ineffective for enhancing visibility within generative engines, while GEO-specific strategies can boost visibility by up to 40\%. However, the effectiveness of these methods varies across sectors, calling for domain-specific research \citep{aggarwal2024}. Additionally, they emphasise the importance of studying proposed GEO optimisation methods in combination, as practitioners are likely to use multiple approaches together, making theoretical research more applicable to real-world practice.

In this report, we aim to contribute to this emerging field by focusing on the \textit{travel and tourism} industry, a content-rich domain that depends heavily on online visibility. Our goal is to explore whether model fine-tuning and domain-adapted optimisation techniques can improve content visibility in generative search settings. We fine-tune a pre-trained BART‑base model on tourism-specific data in a supervised setting, aiming to reduce data and compute requirements - while maintaining high accuracy. We evaluate the performance of optimised website content generated by this model using objective visibility metrics. 

Hence, our contributions are threefold:

\begin{enumerate}
    \item We apply GEO methods on a domain-specific scope: Travel and tourism industry.
    \item We introduce a fine-tuning approach to train an efficient transformer that learns to generate optimised content.
    \item We evaluate the effectiveness of applying multiple GEO methods in combination, reflecting realistic usage scenarios.
\end{enumerate}

The fine-tuned model achieved a ROUGE-L score of 0.249, approximately 25\%, indicating moderate recall of key information. The model achieved a BLEU score of 0.2, indicating 20\% n-gram precision between its output and the reference texts. This suggests a moderate alignment in vocabulary and structure, reflecting a reasonable level of content similarity. The proposed model boosts website visibility by 15.63\% in absolute word count and 30.96\% in position-adjusted word count, with results normalised and outliers excluded to ensure robustness.

The remainder of this report is organised as follows. We first review related work on emerging approaches for generative engine optimisation. Then, we describe our methodology, including data collection, content preprocessing, and the application of domain-specific optimisation as a labelling strategy using existing large language models. We follow with a presentation of our model architecture, before detailing our training methods and evaluation framework. Finally, we present results, discuss and interpret key findings, and conclude with reflections on limitations and future directions.

\section{Related Work}
\label{submission}
Given the novelty of the field and its rapid pace of development, there is limited academic research specifically addressing web content optimisation for generative search engines. Although several scholars have explored the use of generative AI to enhance conventional SEO practices \citep[e.g.][]{ziakis2024, yuniarthe2017}, novel methods going beyond SEO remain sparse.

The emergence of LLM-based search systems gave rise for initial works in this emerging domain. For example, \citet{liu2023} and \citet{menick2022} emphasise verifiability, factual correctness and coherence of LLM-generated responses, but do not address the visibility or ranking of source content within those responses. Sporadic contributions in boosting visibility in generative engines include \citet{kumar2024} who investigate how modifying textual input can enhance product visibility in LLM outputs. However, their work is narrowly scoped to e-commerce scenarios and focuses on content addition rather than transformation.

The foundational contribution to the field concerned - and the root of this report - is the work by \citet{aggarwal2024} who coined the notion of GEO. They propose a black-box optimisation framework designed to enhance the visibility of web content in generative search engine responses. Through the introduction of GEO-bench, a large-scale benchmark encompassing diverse user queries across multiple domains and topics, they demonstrated that GEO-specific strategies could boost content visibility by up to 40\% in generative engine outputs.

\citet{aggarwal2024} introduce seven methods to transform raw website content into a more appealing style to be favored by generative search engines. For instance, these include adopting an authoritative tone, citing credible sources, adding statistical evidence, or improving linguistic fluency. Among these, the most effective methods were statistics addition and citations, as identified through their experimental evaluation. Notably, their approach utilized publicly available GPTs without any model fine-tuning or task-specific adaptation.

To the best of our knowledge, no existing research has explored the application of model fine-tuning to develop domain-tailored models for web content optimisation in generative search engines.

\section{Approach}
\label{submission}
\subsection{Data Generation}
To train our transformer model, we created a custom dataset tailored to the travel and tourism domain. Each data point consists of a pair (\textit{w, w'}), where \textit{w} is raw website content and \textit{w'} is a version optimised for visibility in generative engine responses (label). Since there are no publicly available datasets containing such website content or associated user queries, we generated all data from scratch, following a structured approach informed by the findings of \citet{aggarwal2024}.

The first step of data generation began with creating realistic user queries across 11 subcategories within the domain. To achieve this, we used Meta-Llama-3-8B-Instruct via the HuggingFace API, with a temperature of 0.9 and a set of carefully crafted templates to ensure variation and authenticity. After generation, the query set was cleaned to remove duplicates and ensure diversity.

The second step of the data generation process was conducted to collect raw website content (\textit{w}). For each query, we used the Google Search API to scrape website data from the top five search results. If a website blocked scraping or contained less than 100 characters of content, it was marked as irrelevant for our dataset and thus excluded. We continued scraping subsequent results until five usable sources were collected per query. This gave us a reliable set of raw website texts.

\subsection{Content Pre-Processing}
Raw website content required thorough cleaning before analysis. Due to input length limitations, we trimmed content to a maximum of 4000 characters while preserving essential information. The process began with HTML content extraction using \textit{BeautifulSoup}, which removed script and style elements while maintaining proper text formatting. We focused on eliminating non-essential website elements such as navigation menus, advertisements, footers, and social media widgets that did not contribute substantive information. Quality control metrics ensured only valuable content was included by filtering out websites with insufficient text and avoiding domains known to block scrapers. We also conducted a manual cleaning iteration, to further increase data quality resulting in the database which will be used for labelling - optimising the given text input.

\subsection{Labelling}
To produce optimised versions of raw website content, we transformed each text \textit{w} into a visibility-optimised version \textit{w'} using strategies shown by \citet{aggarwal2024} and treating these labelled outputs as ground truth for our model. Our optimisation pipeline employed Llama-3.3-70B-Instruct-Turbo-Free via the \textit{together.ai} API in a systematic three-step process. (1) Credible-sounding citations were integrated to enhance content authority. (2) Language fluency was improved, complex phrasing was simplified, and key information was preserved. (3) Compelling statistics were placed strategically to reinforce key points. All three phases involved reorganizing content into clearly separable sections with descriptive headings and structured paragraphs.

The pipeline maintained careful pacing between API calls to avoid rate limiting issues. Following optimisation, we conducted thorough data cleaning to remove entries affected by API errors or formatting inconsistencies, ensuring high quality across the dataset. Finally, we obtained a dataset with the raw website text \textit{w} and the respective optimised text \textit{w'} in the final dataset: \textit{X\_final\_filtered\_labelled\_dataset.csv}.

\section{Architecture}
\label{submission}
\subsection{Introduction to BART}
BART (Bidirectional and Auto-Regressive Transformers) is a sequence-to-sequence model developed by Facebook AI in 2019 that combines the strengths of both BERT (Bidirectional Encoder Representations from Transformers) and GPT (Generative Pre-trained Transformer). It was designed as a flexible architecture particularly suitable for a wide range of text generation tasks, including summarisation, paraphrasing, and content rewriting \cite{lewis2019bart}.

For our web content optimisation project, we selected BART as our main model due to its powerful sequence transformation capabilities. The architecture consists of a bidirectional encoder (BERT) and an autoregressive decoder (GPT) \cite{soper2021bart}, making it ideally suited for transforming raw website content into optimised versions - for more visibility in generative search engines.

\subsection{Pre-Training Approach}
Before examining the specific architectural components, it's important to understand BART's distinctive pre-training approach:

\textbf{Denoising Autoencoder:} BART is pre-trained as a denoising autoencoder, using a training objective that corrupts input text and then tasks the model with reconstructing the original. Merging BERT's bidirectional context modelling with GPT's autoregressive generation creates a backbone that adapts efficiently to downstream text‑generation tasks.

\textbf{Noising Functions:} Before any token embeddings or neural layers, one of five different noising functions is randomly chosen and applied to the raw text, producing a "corrupted" version that the model must learn to reconstruct:

\begin{table}[h]
\centering
\begin{tabular}{|p{2.5cm}|p{5cm}|}
\hline
\textbf{Noise type} & \textbf{What it does} \\
\hline
Token Masking & Randomly replace tokens with a special \texttt{[MASK]} token. The model must predict the original token from context. \\
\hline
Token Deletion & Remove random tokens entirely. The model learns to infer which words were dropped. \\
\hline
Text Infilling & Sample spans of text (of varying length) and replace each span with a single \texttt{[MASK]}. The model must generate the missing span. \\
\hline
Sentence Permutation & Split the document into sentences and shuffle their order. The model must recover the original discourse sequence. \\
\hline
Document Rotation & Pick a random token index, rotate the text so that this point becomes the "start." The model must re-align and recover the original start. \\
\hline
\end{tabular}
\caption{BART pre-training noising functions}
\end{table}

This corruption-reconstruction objective makes BART naturally robust to imperfect, noisy, or incomplete input data - a valuable property regarding the variability and messiness prevalent in scraped web content \cite{lewis2019bart}. Research has even shown performance improvements resulting from added noise in input text \cite{soper2021bart}.

\subsection{Tokenisation}
Tokenisation transforms human-readable text into numerical format for neural network processing. BART uses a standard token embedding layer to convert discrete input tokens into dense vector representations. Each token from the vocabulary is mapped to a continuous vector of fixed dimensionality - 768 in the BART-base model \cite{lewis2019bart}. It employs Subword Tokenisation using byte-pair encoding (BPE). Additionally, it maps each token to an integer ID from the vocabulary. Besides this, it creates attention masks, padding, and special tokens. Positional embeddings are then added to these token embeddings to encode the sequential order of the tokens. This allows the model to associate each word with an integer and assign its position within texts and sentences. The resulting combined embeddings are passed into the encoder \cite{alokla2022pseudocode}.

\subsection{Encoder Architecture (6 Layers)}
The encoder in BART ingests the (noised and tokenized) input and produces a rich contextual representation that the decoder will use when generating text. Each encoder block consists of the following layers which perform the following operations:

\begin{enumerate}
\item \textbf{Multi-Head Self-Attention:} This component projects the input into Query ($Q$), Key ($K$), and Value ($V$) spaces which enables it to compute attention using: $$\mathrm{Attention}(Q, K, V) = \mathrm{softmax}(\frac{QK^T}{\sqrt{d_k}})V$$ Variable $d$ resembles the embedding size - being the dimension of the embedding matrix $E \in \mathbb{R}^{V \times d}$, where $d_k$ stands for the dimension per attention head. This means that it allows each token to "look at" every other token and to gather relations within the whole context.

\item \textbf{Residual Connection + LayerNorm:} This adds the attention output back to the original input ("skip connection"), and then normalises it:
$$x \leftarrow \mathrm{LayerNorm}(x + \mathrm{MultiHead}(x))$$

\item \textbf{Position-wise Feed-Forward Network (FFN):} Here two linear transformations with GeLU activation functions are applied in between, where $W$ and $b$ are respectively the weight matrix and the bias term for the first and the second linear transformation within the neural network. The GeLU functions (Gaussian Error Linear Unit) is preferred over the ReLU activation function. The GeLU is partly similar to the general applicable ReLU, but is differentiable everywhere, which helps optimisation and gradient flow in deep models like transformers:
$$\mathrm{FFN}(x) = W_2(\mathrm{GeLU}(W_1 x + b_1)) + b_2$$

\item \textbf{Second Residual Connection + LayerNorm:} This adds the FFN output back to its input and normalises it also once again:
$$x \leftarrow \mathrm{LayerNorm}(x + \mathrm{FFN}(x))$$
\end{enumerate}

\textbf{Encoder Function:} The encoder's self-attention mechanism allows it to weigh the importance of each word or token in the input sequence in relation to all other tokens, enabling the model to capture long-range dependencies and relationships efficiently \cite{alokla2022pseudocode}. The 6-layer depth provides several benefits: (1) Hierarchical feature extraction: Lower layers learn local patterns while deeper layers capture higher-level semantics. (2) Greater modelling capacity: Each additional layer refines the representation. (3) Long-range dependency modelling: The full self-attention mechanism allows BART to consider interactions between any pair of tokens in the sequence regardless of their distance \cite{lewis2019bart}.

\subsection{Decoder Architecture (6 Layers)}
The decoder is responsible for generating the optimised output token-by-token based on the encoded input. Each decoder block consists of the following layers which perform the following operations:

\begin{enumerate}
\item \textbf{Masked Multi-Head Self-Attention:} It allows the decoder to look only at previous tokens it has generated so far, which are \textit{not future tokens}. Additionally it ensures autoregressive decoding (one word at a time) and it uses a lower triangular mask matrix: $$\text{MaskedAttention}(Q, K, V) = \mathrm{softmax}(\frac{QK^T}{\sqrt{d_k}} + \text{mask})V$$

\item \textbf{Encoder–Decoder Cross-Attention:} This takes the encoder's output as keys and values and the decoder's hidden states as queries and enables the decoder to focus on relevant parts of the input sequence.

\item \textbf{Feed-Forward Network (FFN):} This neural network is the basic building block and is identical to the encoder's FFN in applying a non-linear transformation independently to each token's representation.

\item \textbf{Residual Connections + Layer normalisation:}
Each sub-layer applies a residual connection by adding its output to its own input, followed by layer normalisation. Residual connections mainly help mitigate the vanishing gradient problem. Note that this input already includes the positional embeddings (added once at the start), but positional encodings are not repeatedly added in residual connections.
\end{enumerate}

\textbf{Decoder Function:} The decoder generates output sequences autoregressively, which means it is relying on previously generated tokens while attending to the encoder's output \cite{tan2024qarrfsqa}. This cross-attention mechanism enables the decoder to: (1) It uses the encoder’s full, two‑way grasp of the input to guide its writing. (2) As it picks each word, it zooms in on the most relevant parts of that encoded input. (3) It keeps its own previous words in mind so the sentence stays coherent.

\subsection{Complete BART Model Architecture}
\begin{figure}
    \centering
    \includegraphics[width=0.5\linewidth]{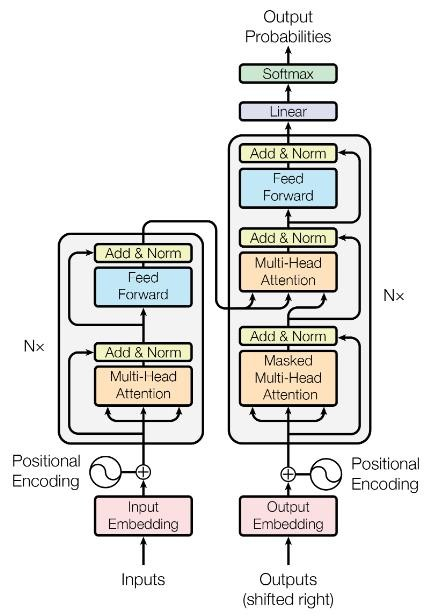}
    \caption{Outline of BART's architecture. Source: \cite{vaswani2017}}
    \label{fig:enter-label}
\end{figure}

The full BART architecture integrates the encoder and decoder into a cohesive sequence-to-sequence model with three crucial components:

\textbf{Model Components:}

\begin{enumerate}
\item \textbf{Shared Token Embeddings:} BART uses a single shared embedding layer for both encoder and decoder, whereas the parameter sharing reduces total parameter count and encourages consistency.

\item \textbf{Encoder-Decoder Pipeline:} The encoder transforms input text into contextualized representations. After that, the decoder uses these representations in its cross-attention layers. This cross-attention acts as an information bridge between encoding and generation.

\item \textbf{Language Modelling Head:} Finally, the decoder’s hidden states are mapped through a linear layer to produce a probability distribution over the vocabulary. This output layer shares weights with the input‑embedding layer.
\end{enumerate}

\textbf{Parameter Configuration:}

\begin{table}[h]
\centering
\begin{tabular}{|p{2.5cm}|p{5cm}|}
\hline
\textbf{Parameter} & \textbf{Configuration} \\
\hline
Vocabulary size & Derived from the pre-trained tokenizer \\
\hline
Position embeddings & Maximum length of 1024 tokens \\
\hline
Model depth & 6 layers each for encoder and decoder \\
\hline
Model width & 768-dimensional embeddings with 12 attention heads \\
\hline
Feed-forward networks & 3072-dimensional inner representations (4× the model dimension) \\
\hline
Regularisation & Dropout rate of 0.1 \\
\hline
Initialisation & Parameters sampled from N(0, 0.02) \\
\hline
\end{tabular}
\caption{BART-base model parametrisation}
\label{tab:bart-params}
\end{table}

The parametrisation of BART-base is balanced for both expressiveness and computational efficiency. With six layers each in the encoder and decoder, hidden layers with size of 768 dimensions, and 12 attention heads, the model offers a strong balance between expressiveness and computational tractability.

\subsection{Why choosing pre-trained BART instead of training from scratch}
Our project initially explored implementing BART from scratch as an educational exercise, but we encountered several practical limitations that led us to adopt the pre-trained BART-base model with fine-tuned parameters instead:

\textbf{Data Scarcity Barrier:} The modest performance metrics we observed stemmed primarily from our limited dataset size (only 1905 texts). The original (pre-trained, ready to be imported) BART was trained on over 160GB of text data including books, news, stories, and web text \citep{aggarwal2024}. At the same time, our from-scratch implementation is trained on 1905 website content examples, which shows the big discrepancy in data usage size. This represented a fundamental data scale mismatch of several orders of magnitude. Transformers require extensive exposure to linguistic patterns to develop robust internal representations.

\textbf{Parameter Initialisation Challenge:} When training from scratch, our model began with randomly initialised parameters (following the N(0, 0.02) distribution). These random starting points required substantial data and training iterations to converge to useful representations, whereas pre-trained models begin with parameters that already encode rich linguistic knowledge.

\textbf{Computational Efficiency Trade-off:} The computational demands we encountered (1.5 hours for 3 epochs on a 16GB RAM Apple MacBook M1 Pro) highlighted the resource intensity of training transformer models. This practical constraint limited our ability to train for sufficient epochs to reach optimal performance. Training on Google Colab crashed due to usage limitations.

\subsection{BART's Advantages for Our Content optimisation}
BART's architecture makes it particularly well-suited for our web content transformation task. BART's Transformer architecture allows direct paths for gradients and information to flow between all tokens suitable for long website content with distributed information. Bidirectional context (BERT) and autoregressive Generation (GPT) excels at comprehending content and transforming it

While building BART from scratch provided valuable architectural understanding, for the task concerned, leveraging a pre-trained BART model and fine-tuning it on our website content optimisation task yields superior performance while requiring significantly less data and computational resources. For this reason, we chose the pre-trained BART model. The full model architecture of BART-base can be found in the notebook \textit{/2 Deep Learning Model/BART from scratch/OurBART.ipynb}, while the pre-trained model used for the project can be found in the \textit{/2 Deep Learning Model/Code/BART\_base\_supervised\_model.ipynb}. The \textit{OurBART.ipynb} notebook acts as a "zoom-in" into the imported model, which is reflected as Step 3 in the pre-trained \textit{BART\_base\_supervised\_model.ipynb} file.

\section{Training Methods}
\label{sec:training_methods}
\subsection{Tokenisation of the dataset}
The generated dataset contains travel-related web pages (\textit{w} and \textit{w'}) that have slightly different length. To preserve the natural imbalance between short and long content while keeping training and validation distributions aligned, the raw page length is first partitioned into ten equiprobable bins. Within every bin the same proportion of items is sampled, yielding splits of $80\,\%$ for training, $20\,\%$ for validation and a disjoint test set composed of 50 unseen queries (250 instances of websites).

\subsection{Training-argument rationale}
\label{sec:train_args}
\textbf{Optimiser and learning-rate schedule:} Fine-tuning uses the AdamW optimiser \citep{kingma2015adam,loshchilov2019decoupled} with a target learning rate of $3\!\times\!10^{-5}$ and decoupled weight decay for the regularisation in training ($\lambda\!=\!0.01$). To escape shallow minima the learning rate increases from near zero for the first 250 updates reaching the target learning rate of $3\!\times\!10^{-5}$. The learning rate then follows a cosine decay towards zero, a heuristic shown to favour both faster convergence and better final generalisation \citep{loshchilov2017sgdr}.

\textbf{Controlling over-fitting:} Besides regularizing with weight decay, the transformer already embeds dropout at $0.1$. Rather than intervening with early stopping, the run stores a checkpoint at the end of every epoch; the model with the smallest validation loss is retained. This strategy avoids premature termination caused by high-variance sequence metrics while still protecting against over-fitting.

\textbf{Handling overly short and repetitive outputs:} During early runs, we observed that the model often produced outputs that were much shorter than the ground truth. This was problematic because short outputs could still score high on ROUGE-L by simply copying a few keywords, yet they failed to deliver the full rewritten version desired. To address this, two constraints were applied during decoding: We introduce a length penalty \(\alpha = 1.1\) to the model’s scoring so that shorter outputs incur a larger penalty, making longer responses more likely. And we also activated a \texttt{no\_repeat\_ngram\_size} of 3, a rule that forbids the same 3-word phrase from appearing more than once, which helps reduce exact copying.

\section {Numerical results}
\label{sec:numerical_results}
\subsection{Experimental settings}
The hyperparameters of proposed model (our fine-tuned model) and the BART-base (from Facebook AI) differ in the size of the context window that flows through the encoder–decoder stack. Our proposed model enlarges the encoder window to \texttt{MAX\_IN}$=384$ and caps the decoder at the same length, while BART-base is set to \texttt{MAX\_IN}$=256$ and \texttt{MAX\_OUT}$=448$ tokens. All remaining hyper-parameters, optimisation schedule and hardware budget are held constant.

\subsection{Optimisation objective}
\begin{figure}[h]
  \centering
  \includegraphics[width=\linewidth]{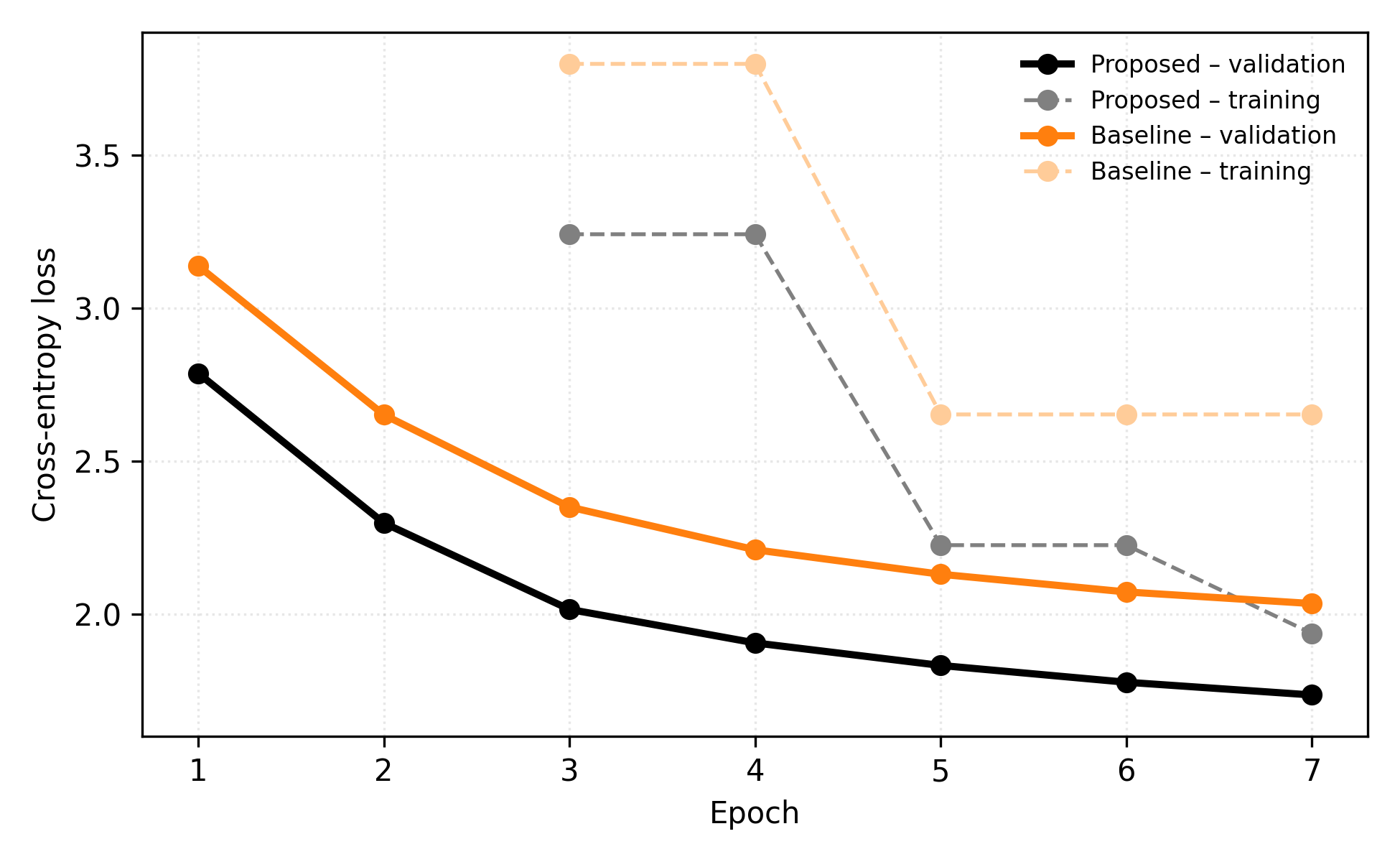}
  \vspace*{-1ex}
  \caption{Training and validation loss for Proposed and Baseline models.}
  \label{fig:loss-curves}
\end{figure}

The optimisation objective during fine-tuning is the minimisation of token-level cross-entropy, which quantifies the divergence between the true next-token distribution \(q\) and the model’s predicted distribution \(p_{\theta}\). For a target sequence \(\mathbf{y} = (y_{1}, \dots, y_{T})\) generated conditionally on an input \(\mathbf{x}\), the cross-entropy is defined as:

$$\mathcal{H}(q, p_{\theta}) = - \sum_{t=1}^{T} q(y_{t}) \log p_{\theta}(y_{t} \mid \mathbf{x}, y_{<t})$$

which reduces to the negative log-likelihood of the observed tokens under a one-hot encoding of \( q \). This quantity is computed at each training step and averaged between tokens in the batch. Figure~\ref{fig:loss-curves} shows that the proposed model not only starts better, but also converges faster in the loss of validation, despite its larger input window.

\subsection{Evaluation metrics}
To evaluate the generative capabilities of the model beyond token-level likelihoods, four complementary metrics are calculated after every epoch:

\begin{enumerate}
  \item \textbf{ROUGE-L}: Length of the longest shared word sequence between generated output and reference. Higher values mean better coverage of key phrases and structure.
  \item \textbf{BLEU}: Precision of matching 1–4‑word sequences (clipped $n$-gram precision ($n\!\le\!4$)), with a penalty if the output is too short (brevity penalty). Higher scores indicate more exact overlaps and fewer invented phrases.
  \item \textbf{Length ratio}: Ratio of output length to reference length. A value near 1 means the lengths match.
  \item \textbf{Perplexity (PPL)}: Model’s average uncertainty per token. Lower PPL means more fluent, predictable text.
\end{enumerate}

\begin{figure}[h]
  \centering
  \includegraphics[width=\linewidth]{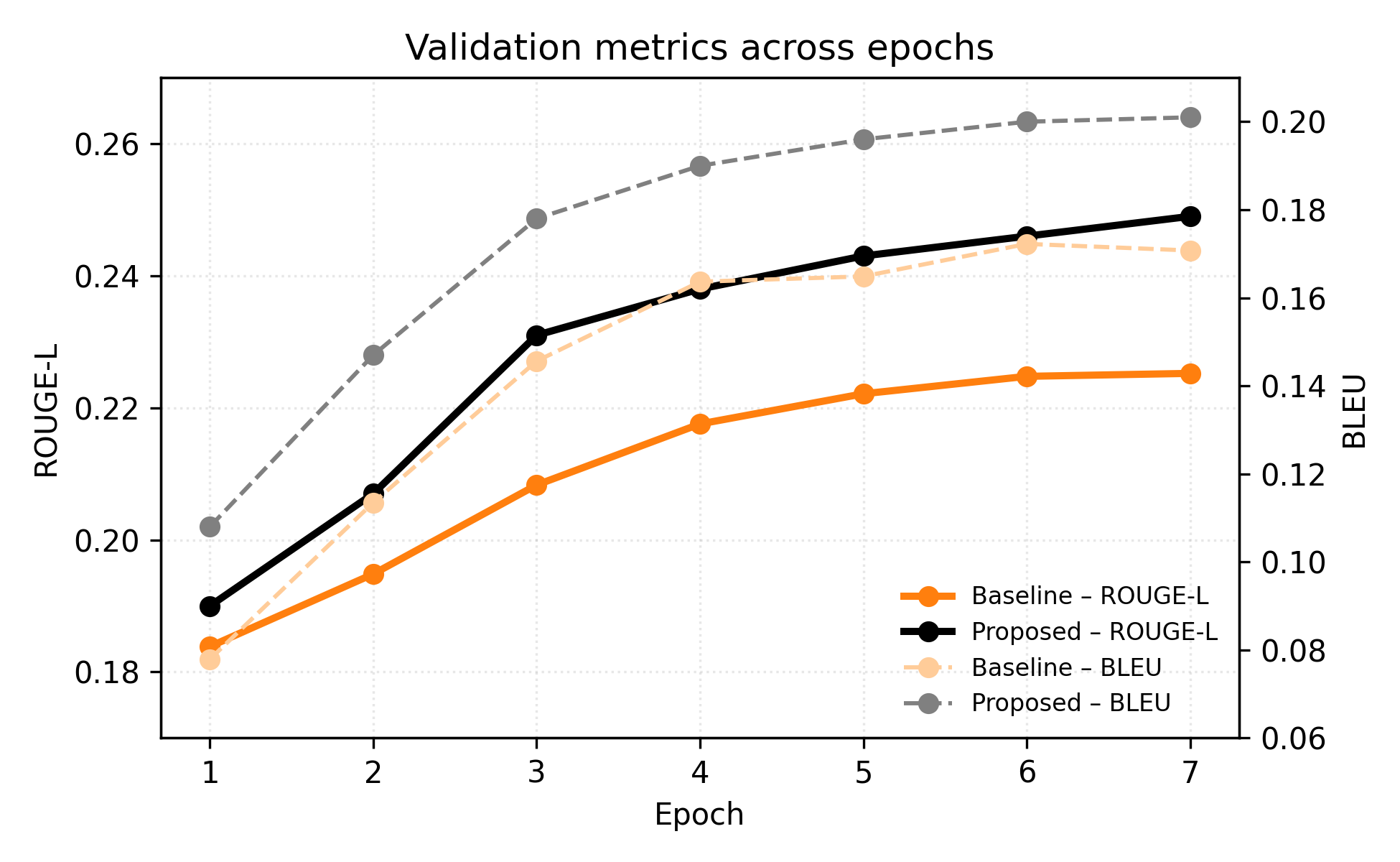}
  \vspace*{-1ex}
  \caption{Validation \textsc{ROUGE-L} (solid) and
           \textsc{BLEU} (dashed) for Proposed and Baseline models.}
  \label{fig:metric-curves}
\end{figure}

Figure~\ref{fig:metric-curves} contrasts \textsc{ROUGE-L} and \textsc{BLEU}. Both models plateau after four epochs, and consistent improvements are observed for our proposed model.

\subsection{Final test performance}
Table~\ref{tab:test-metrics} summarises the scores on the 250-instance test set (50 queries~$\times$~5 documents each).

\begin{table}[h]
\centering
\caption{Best checkpoint (\emph{epoch~7}) on the test set.}
\label{tab:test-metrics}
\vspace{.5em}
\begin{small}
\begin{tabular}{@{}lcccc@{}}
\toprule
\textbf{Model} & \textbf{ROUGE-L} $\uparrow$ & \textbf{BLEU} $\uparrow$
              & \textbf{PPL} $\downarrow$ & \textbf{Len. ratio} $\approx$1 \\
\midrule
Baseline & 0.226 & 0.173 & 1.71 & 1.01 \\
Proposed & \textbf{0.249} & \textbf{0.200} & \textbf{1.50} & 1.00 \\
\bottomrule
\end{tabular}
\end{small}
\vspace*{-1em}
\end{table}

\section{Interpretation}
\label{sec:interpretation}

\subsection{Model Evaluation}
\textbf{Loss curve interpretation (Figure ~\ref{fig:loss-curves}):} Both models show the expected monotonic decline in cross-entropy as training progresses, but the proposed configuration starts lower and falls steeper, converging to a validation loss below~2.0 by epoch~7. In contrast, the baseline flattens near~2.1. The gap suggests that the larger encoder window (384 vs.\ 256~tokens) helps the decoder ground its predictions in a richer context, shortening the "learning runway". Training loss sits above validation loss for both models because the training split contains systematically longer passages whose input side is artificially truncated. The parallel trajectories and the absence of sudden divergence argue against over-fitting.

\textbf{ROUGE-L and BLEU interpretation (Figure ~\ref{fig:metric-curves}):} Quality metrics echo the loss picture. BLEU and ROUGE-L rise sharply during the first three epochs, plateau around epoch~5, and stabilise thereafter showing that the optimiser has reached a region of diminishing returns rather than oscillating. Across the entire sweep the proposed model maintains a consistent $\approx$0.02 absolute lead in ROUGE-L and a $\approx$0.03 lead in BLEU, validating that its lower loss does translate into better content overlap and \(n\)-gram precision. The fact that BLEU (dashed) improves in lock-step with ROUGE-L (solid) indicates that the gains are not merely caused by copying longer substrings but by more faithful phrasing at finer granularity.

\textbf{Length control efficacy:} Prior runs produced \(\approx\!200\)-token summaries of inputs that artificially inflated ROUGE-L.  After adding a \(384\)-token ceiling, a mild length penalty (\(\alpha=1.1\)) and “no-repeat 3-gram’’ blocking, the system delivers outputs whose word counts track the reference within five percent, while still gaining metric points.

\subsection{GEO Evaluation}
To evaluate the effectiveness of our optimisation strategies, we conducted a second-stage model assessment focused on website visibility within generative responses. Using our final test dataset, we employed our fine-tuned BART model to transform unoptimised website content \textit{w} into its optimised counterpart \textit{w'}. A subset of 50 evaluation queries was created from the original set of 250, with each query containing five unoptimised websites and one randomly selected optimised version.
We queried Llama-3.3-70B via API in two conditions: (1) using only the five unoptimised sources and (2) randomly replacing one with its optimised version. In both scenarios Llama was asked to answer the query provided only by using information out of these sources and citing them numerically. Inspired by \cite{aggarwal2024}, we assessed visibility changes in the Llama response from (1) to (2) based on two metrics: absolute word count improvement and position-adjusted word count improvement. The absolute word count $wc$ is the number of words of the source $c_{i}$ in the sentences $s$ of the response $r$ citing $c_{i}$, mathematically defined as 

$$\mathrm{wc}_{}(c_i, r) = \sum_{s \in S_{c_i}} |s|$$

The position-adjusted wordcount is the same metric, but weighted by a linear decaying function of the citation position. Citations that appear earlier in a generative response tend to receive greater user attention, so the second metric places more emphasis on these positions \citep{aggarwal2024}.

$$\mathrm{wc}_{\text{adj}}(c_i, r) = \sum_{s \in S_{c_i}} |s| \cdot \left(1 - \frac{\mathrm{pos}(s)}{|S|} \right)$$

To ensure robustness, we normalised the results and excluded outliers. Our results show that the model improves website visibility by \textbf{15.63\%} based on absolute word count and by \textbf{30.96\%} using the position-adjusted word count metric. This is consistent with results of \cite{aggarwal2024}, but faces several limitations. The sample size (n=50) is too small for statistical significance, Llama’s token constraints restricted context and response length leading to extreme values, and noisy data occasionally introduced bias (e.g. poor website-query match).

\section{Conclusion}
\label{sec:conclusion}
This work delivers the first end-to-end demonstration that a medium-size sequence-to-sequence transformer, when lightly fine-tuned on synthetic optimisation pairs, can measurably increase a travel website’s visibility inside LLM-generated answers. Relative to a length-matched baseline, our GEO-tailored model lifts ROUGE-L by \textbf{+2.3\,pp}, BLEU by \textbf{+2.7\,pp} and reduces perplexity by \textbf{12\,\%}, all while holding the output length within the 5\,\% corridor. A downstream evaluation with Llama-3.3-70B confirms that these textual improvements translate into a \textbf{+30.96\,\%} boost in position-adjusted word count, bringing the rewritten pages noticeably closer to the “top answer” zone in generative search. Crucially, these gains are obtained without retraining a multi-billion-parameter backbone, without access to proprietary retrieval infrastructure, and on a single commodity GPU - conditions that mirror the realities of most content publishers. The study therefore provides concrete evidence that GEO is tractable with open-source tooling and modest compute, offering a practical bridge between academic research and day-to-day digital-marketing needs.

However, this work was significantly constrained by limited computational resources, which influenced model selection and training efficiency. Additionally, the novelty of the task introduced further challenges, particularly the absence of existing labelled datasets suitable for fine-tuning. Synthetically constructing a high-quality, domain-specific training dataset at sufficient scale proved non-trivial. Despite systematic data generation efforts, issues such as noise, formatting inconsistencies, and weak alignment between website content and user queries emerged as limiting factors. These imperfections had a measurable impact on model performance during evaluation, particularly in cases where the relevance of query-content was poor and diminished visibility outcomes.

Future research can address these limitations by scaling dataset generation using more robust scraping and alignment techniques, and evaluating models across other domains. Additionally, reinforcement learning could directly optimize for visibility metrics, improving practical performance in generative search settings. Lastly, future research should explore retrieval-augmented generation (RAG) to improve factual grounding in optimised web content.

Overall, this study demonstrates that targeted fine-tuning with domain-specific strategies can meaningfully enhance web content visibility in generative search, even under constrained resources.

\section {Statement about individual contributions}
\label{statement}
All four group members have worked equally on this project and provided equal parts of work. Two group members have focused on the data generation and preparation side while the two others have focused on researching the right model to chose and to work with. In the end, tasks considering model creation, fine-tuning, and evaluation have been worked on by all four members interchangeably.







\end{document}